\documentclass[compsoc]{IEEEtran}
\usepackage[authoryear, round]{natbib}
\setcitestyle{aysep={}}
\usepackage{amsmath,amssymb,amsfonts,float}
\newcommand\numberthis{\addtocounter{equation}{1}\tag{\theequation}}
\usepackage{algorithmic}
\usepackage{graphicx}
\usepackage{textcomp}
\usepackage{xcolor}
\usepackage[inline]{enumitem}
\usepackage{sidecap}
\sidecaptionvpos{figure}{t}
\usepackage{etoolbox}
\usepackage[hidelinks]{hyperref}
\usepackage{academicons}
\usepackage{xcolor}
\usepackage[labelfont=bf, 
            font=footnotesize,
            labelsep=space]{caption}
\makeatletter
\patchcmd{\@makecaption}
  {\scshape}
  {}
  {}
  {}
\makeatother

\newcommand{\orcid}[1]{\href{https://orcid.org/#1}{\includegraphics[width=10pt]{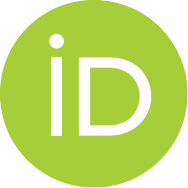}}}

\def\BibTeX{{\rm B\kern-.05em{\sc i\kern-.025em b}\kern-.08em
    T\kern-.1667em\lower.7ex\hbox{E}\kern-.125emX}}

\begin{document}

\title{Bio-Inspired, Task-Free Continual Learning \protect\\ through Activity Regularization
\thanks{This article is published as part of the Special Issue on 'What can Computer Vision learn from Visual Neuroscience?'}
}

\author{\IEEEauthorblockN{Francesco Lässig$^1$ \orcid{0000-0001-9535-9849}, Pau Vilimelis Aceituno$^1$ \orcid{0000-0002-1218-4009}, Martino Sorbaro$^{1,2}$ \orcid{0000-0002-0182-7443}, Benjamin F. Grewe$^1$ \orcid{0000-0001-8560-2120}}

\IEEEauthorblockA{\textit{\textsuperscript{1}Institute of Neuroinformatics, University of Zürich and ETH Zürich, Switzerland \\
\textsuperscript{2}AI Center, ETH Z\"urich, Switzerland} \\
flaessig@ethz.ch, bgrewe@ethz.ch}
}

\maketitle

\begin{abstract}
The ability to sequentially learn multiple tasks without forgetting is a key skill of biological brains, whereas it represents a major challenge to the field of deep learning. To avoid catastrophic forgetting, various continual learning (CL) approaches have been devised. However, these usually require discrete task boundaries. This requirement seems biologically implausible and often limits the application of CL methods in the real world where tasks are not always well defined. Here, we take inspiration from neuroscience, where sparse, non-overlapping neuronal representations have been suggested to prevent catastrophic forgetting. As in the brain, we argue that these sparse representations should be chosen on the basis of feed forward (stimulus-specific) as well as top-down (context-specific) information. To implement such selective sparsity, we use a bio-plausible form of hierarchical credit assignment known as Deep Feedback Control (DFC) and combine it with a winner-take-all sparsity mechanism. In addition to sparsity, we introduce lateral recurrent connections within each layer to further protect previously learned representations. We evaluate the new sparse-recurrent version of DFC on the split-MNIST computer vision benchmark and show that only the combination of sparsity and intra-layer recurrent connections improves CL performance with respect to standard backpropagation. Our method achieves similar performance to well-known CL methods, such as Elastic Weight Consolidation and Synaptic Intelligence, without requiring information about task boundaries. Overall, we showcase the idea of adopting computational principles from the brain to derive new, task-free learning  algorithms for CL.
\end{abstract}

\begin{IEEEkeywords}
continual learning,
bio-inspired,
sparsity,
feedback,
lateral inhibition,
activity regularization
\end{IEEEkeywords}

\section{Introduction}
The mammalian brain has an astonishing ability to continually form new memories while preserving previous ones. In contrast, artificial neural networks (ANNs) are prone to \textit{catastrophic forgetting} (CF) when trained on a sequence of tasks or datasets \citep{mccloskey1989catastrophic}. 

For deep ANNs, a range of continual learning (CL) approaches have been devised that include modifications to the network architecture, loss function, or the implicit or explicit storage of previous task data \citep{vandeven2019}. Usually, these methods require external information about a task switch. This is in stark contrast to natural environments, where tasks are usually not well defined and need to be inferred from context.

To address the CL problem, brain-inspired approaches have been developed \citep{kudithipudi2022biological, parisi2019continual}. For example, \cite{French1993} pointed out that the problem of CF might not be intrinsic to biological neural networks, but is rather an effect of distributed and overlapping task representations that emerge when using the standard backpropagation (BP) algorithm. In line with this idea, it has been suggested that biological networks might avoid CF by representing information through a sparse, but task-specific subset of neurons and synapses to which learning is restricted \citep{lin2014sparse,manneschi2021sparce,French1993}. Other approaches relax the idea of restricting learning to sub-populations to the more general notion of learning within restricted subspaces \citep{duncker2020organizing}.

In this work we exploit the idea of restricting learning to task-specific, sparse representations with the goal to derive a novel, bio-inspired task-free CL method. In line with the pervasive recurrence observed in the visual cortex \citep{van2020going}, we argue that a task-specific sparsity mechanism should not only incorporate feedforward information (bottom-up) coming from lower hierarchical layers but also error feedback information coming from higher areas (top-down). To render both forms of information usable for such informed sparsity, we adopt Deep Feedback Control (DFC), a bio-plausible deep learning framework in which every neuron integrates inputs from the previous layer, as well as top-down error feedback during learning. To enforce sparsity, we combine DFC with a winner-take-all (WTA) mechanism and restrict learning of the feedforward weights to active neurons. To stabilize and protect previously learned representations, we further introduce intra-layer recurrent weights that are updated through a Hebbian-type learning rule. In the following, we term this new, combined method \textit{sparse-recurrent} DFC.

To explain the basics of our algorithm, we first present related work in Section \ref{sec:background}. Then, in Section~\ref{sec:setup}, we provide implementation details on how we modified the DFC learning dynamics to integrate the two major factors required for CL --- sparsity and intra-layer recurrent connections. In Section \ref{sec:results}, we show that the introduction of these additional bio-plausible elements helps to stabilize learning and to reduce forgetting by regularizing neural activity.
We compare our approach with other established regularization-based CL methods, and show that sparse-recurrent DFC performs comparably well despite completely lacking information on task boundaries. Finally, we analyze the resulting task representations in order to better understand the mechanisms behind the observed improvement in CL performance.

\begin{figure}
    \centering
\includegraphics[width=84 mm]{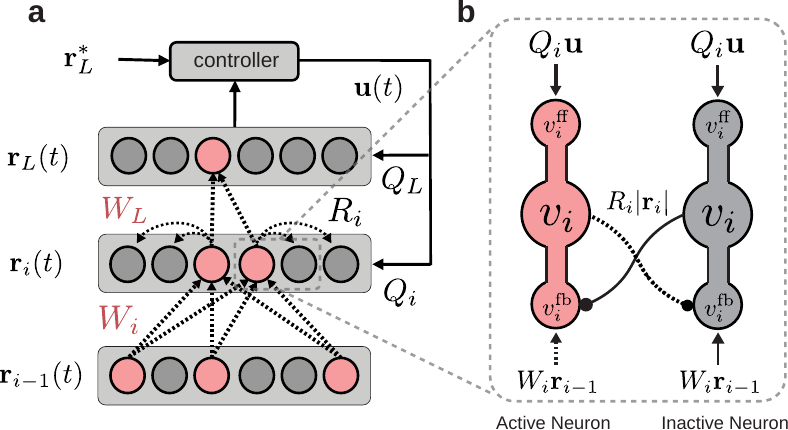}
    \caption{
    \textbf{a} Schematic of the sparse-recurrent DFC network and its top-down feedback controller. The $r_i(t)$ values denote neuron activation vectors for layer $i$, whereas $r_L^*$ represents the desired network output. Learning is based on a dynamic process during which neurons integrate feedforward and feedback signals until the network converges to a sparse target representation minimizing the loss. Weight updates (dashed lines) of forward weights $W_i$ are restricted to neurons that are active at convergence (red). Lateral recurrent weights $R_i$ into inactive neurons are updated via a Hebbian-like learning rule. The $Q_i$ values denote feedback weights, and $u(t)$ refers to the control signal.
    \textbf{b} Detailed zoom into layer $i$ showing one active (pink) and one suppressed (grey) neuron. $v_i^\mathrm{ff}$, $v_i$ and $v_i^\mathrm{fb}$ represent feedforward, feedback and combined activity, respectively. The solid lines represent weights that will not be changed, whereas dashed lines show weights which will be updated} 
    
    \label{fig1}
\end{figure}

\section{Background}\label{sec:background}
\subsection{Computational strategies for continual learning} 
To overcome catastrophic forgetting, researchers developed a variety of different strategies that can roughly be classified into three categories:

1) \textit{Replay methods} rely on implicitly or explicitly storing and revisiting previous data while learning new tasks. This can be accomplished by storing small subsets of previously seen data in a memory buffer, or by training a generative model \citep{shin2017continual}. However, we do not consider data replay in this work, since we are interested in methods based on bio-plausible plasticity, without relying on external data storage.

2) \textit{Regularization methods} constrain learning to preserve parameters that are important for previous tasks, usually by adding specialized loss terms. Elastic Weight Consolidation (EWC) and Synaptic Intelligence (SI) are commonly used representatives of this family, which we adopt as comparison benchmarks. In EWC \citep{Kirkpatrick2016}, after the network converges on a task, the Fisher information of the first task's loss is computed through a sampling mechanism. The Fisher term contains information on parameter importance relative to the first loss, and is added as a regularization term to the loss for the following task. Synaptic Intelligence \citep{Zenke2017} works through a similar mechanism, but parameter importance is estimated online based on how much of the decrease in loss can be attributed to the variation of each given parameter. In both cases, the regularization term is added to the loss at the end of each task, and information on task boundaries is therefore required.

3) \textit{Architectural methods} are based on structural changes such as freezing weights, or adding and removing neurons \citep{rusu2016progressive}. Alternatively, neurons can be dynamically gated based on context \citep{masse2018alleviating, oshg2019hypercl}. Context, however, is usually externally provided rather than inferred by the network itself, which is a strong assumption that may not always hold for real world scenarios. In another approach, a dedicated system, inspired by the role of the prefrontal cortex, is used to detect contextual information instead \citep{zeng2019continual}. In this work, we adopt a similar gating-based approach, in which, conversely, gating is provided by recurrent activity independently of external task information.

\subsection{Continual learning in the brain}
Although CL in the brain is not well understood, it is likely that various mechanisms are at play simultaneously, with some being loosely connected to the three CL strategies described above \citep{kudithipudi2022biological}. 

In neuroscience, the trade-off between fast learning and slow forgetting is known as the stability-plasticity dilemma. To avoid this issue, the interaction between a more plastic system, the hippocampus, and a more stable system, the neocortex, has been suggested as a long term memory storage mechanism, akin to a data replay strategy \citep{van2020brain}. On the other hand, biological networks might control the stability/plasticity of individual synapses through mechanisms collectively referred to as \textit{metaplasticity}.
Through metaplasticity, synapses that are particularly important for solving previously learned tasks are left unaltered when learning new tasks, while less relevant synapses are made available to store new information, analogously to certain regularization-based approaches in CL \citep{jedlicka2022contributions}. Next, \textit{neurogenesis}, the birth of new neurons, is sometimes considered equivalent to architectural approaches that gradually grow the networks when new tasks are learned. However, the number of newly generated neurons is often insignificant in biology. It is therefore contested whether neurogenesis plays a role in CL \citep{parisi2018role}. 

Finally, animal brains heavily rely on \textit{context} to flexibly switch between tasks and to direct learning to task-specific neurons and synapses. For example, previous studies have shown that afferents of the olfactory nucleus in rats provide contextual input from other brain areas, thereby enabling dynamic and flexible task learning \citep{levinson2020context}. This not only enables context-specific gating of neuronal responses to the same stimulus for different environments or tasks but it also facilitates forward-generalization. Similarly, the release of specific neuromodulators (e.g. dopamine) has been linked to the gating of activity and to learning based on context  \citep{kudithipudi2022biological}. Overall, it is likely that in biological networks the modulation of neuronal activities, either through hierarchical top-down feedback or specific neuromodulators, directs learning to the most salient aspects of the task, while protecting older memories that are irrelevant in the current context.

\subsection{Task-free continual learning} 
Van de Ven and Tolias (\citeyear{vandeven2019}) defined three CL scenarios for which training is organized sequentially on each task and performance is evaluated as the average accuracy on all previously learned tasks: \begin{enumerate*}
    \item in \textit{task-incremental learning} (task-IL), the task ID is available during training and at test time; \item in \textit{domain-IL}, the task ID is available during training but not at test time;
    \item in \textit{class-IL}, the task ID is available during training, but at test time the model must report the task ID alongside solving the task.
\end{enumerate*}

In all these scenarios, however, information on \textit{task boundaries} is provided during training, i.e.~the model knows when training on one task $i$ ends and training on a new task $i+1$ begins. Most CL strategies need this information to update the loss or the network structure in preparation for the new task. However, such discrete changes in the loss or network structure do not seem biologically plausible. Therefore, in this paper, we focus on domain-IL, and on the more challenging class-IL, but in a setting where task information is entirely omitted during both training testing.

This so-called \textit{task-free} form of continual learning is generally less studied, although a few examples have appeared in recent years. The majority of these follow a data storage and replay paradigm \citep{aljundi2019gradient, wang2022improving, rao2019continual}, which we do not consider in this work. \cite{lee2020neural} adopt an architectural approach, based on an expanding set of experts which, in turn, deal with new tasks. Among regularization-based methods, \cite{Laborieux2021} propose a metaplasticity-inspired mechanism, but so far limited to feedforward, binary networks. \cite{aljundi2019task} circumvent the problem of task boundaries by heuristically detecting plateaus in the evolution of the loss, which signal the end of learning for a task, and use a mixed replay and regularization strategy. Finally, \cite{pourcel2022online} mix an architectural method with replay using a dynamic content-addressable memory for online class-IL.

To clarify how our method fits into this landscape of brain-inspired algorithms, we next provide details on our CL approach, which combines DFC, sparsity, and recurrent Hebbian-like connections.

\section{Activity regularization through sparsity and recurrent gating}
\label{sec:setup}

\subsection{Learning dynamics}
During training, the neuronal dynamics within the DFC network~\citep{Meulemans2022} can be described by a differential equation that takes into account the feedforward inputs $v_{i}^{\mathrm{ff}}$ as well as the feedback control signal $v_{i}^{\mathrm{fb}}$ according to

\begin{equation}
\label{eq:network_dynamics}
\begin{aligned}
    \tau_v \dot{ v}_i(t) &= - v_i(t) + \:\:\: \:\:\:\:\:\:\: v_{i}^{\mathrm{ff}}(t) \:\:\:\:\:\, +v_{i}^{\mathrm{fb}}(t)\\ 
    &= - v_i(t) + W_i \phi\big(v_{i-1}(t)\big) + Q_iu(t)
\end{aligned}
\end{equation}
where the pre-non-linearity neuron activations in layer $i$ at time $t$ are denoted by $v_i(t)$, and the incoming weights by $W_i$. $\phi$ refers to the activation function while the neuron output is given by $r_i=\phi(v_i(t))$. The feedback signal $u(t)$ is computed as a function of the network output error as described by \cite{Meulemans2022}. $u(t)$ is then fed back to each neuron of the network via the feedback weights $Q_i$. During learning, the feedforward network and the feedback controller constitute a recurrent dynamical system that converges to a final target at which the neuron activations $v_{i,\mathrm{ss}}$ minimize the output error and stabilize the feedback signal $u(t)$.

The forward weights are learned by comparing each neuron's target activation $r_{i,\mathrm{ss}}$ to its feedforward-driven activation $\phi(v_{i,\mathrm{ss}}^{\mathrm{ff}})$ upon converging to the stable state (ss):

\begin{equation}\label{eq:update_W_ss}
    \Delta W_{i} = \eta   (r_{i,\mathrm{ss}} - \phi(v_{i,\mathrm{ss}}^{\mathrm{ff}}))
    r_{i-1,\mathrm{ss}}^T
\end{equation}
where $r_{i-1,\mathrm{ss}}$ is the pre-synaptic activity with controller feedback, $r_{i,\mathrm{ss}}$ is the activity of the neuron with feedback and $\phi(v_{i,\mathrm{ss}}^{\mathrm{ff}})$ is the postsynaptic neuron activity without feedback. In sparse-recurrent DFC, we additionally center each weight update to have zero mean before applying it. This is done in order to prevent a small group of neurons to be more excitable and dominate the winner-take-all mechanism described in the next subsection. The feedback weights $Q_i$ can be learned~\citep{Meulemans2021,Meulemans2022}, but we simplify the learning of the feedback pathway and re-initialize $Q_i$ as the Jacobian of the loss with respect to the neuron activations for every data point.

The update rule from Equation \ref{eq:update_W_ss} implements a learning paradigm where weight updates are determined by neural activity. This opens the possibility of regularizing weight updates indirectly by modulating neural activity. We will refer to this strategy as activity regularization. In the next sections, we describe how activity regularization (e.g. sparsity) can be utilized to reduce interfering weight updates between representations of different inputs belonging to different tasks.

\subsection{Dynamic sparsity}
To gradually modulate the network activations towards sparse, non-overlapping representations, we add a winner-take-all (WTA) mechanism on top of the existing DFC network. At each time step $t$, we set a small fraction $s_{i}(t)$ of neurons to be zero. We then increase $s_{i}(t)$ dynamically until the desired sparsity for the stable state $s_{i, \mathrm{ss}}$, which is a hyperparameter fixed for each layer $i$. We refer to these hyperparameters as sparsity levels.
Although this mechanism is sufficient to reach desired sparsity levels, the network cannot learn to suppress specific neurons because forward connections to inactivated neurons are frozen. To alleviate this problem, we introduce an additional set of connections with the aim of learning which neurons are allowed to fire together, and which neurons are mutually exclusive. This way, we provide a way for the network to stabilize and protect the neuron populations that constitute specific representations.

\subsection{Gating neuron activity through lateral recurrent connections}
We stabilize neuron populations involved in learned representations by introducing lateral recurrent connections. Because we want to strongly influence which neurons are active, we implement lateral connections with a gating effect that multiplies activations by a factor between 0 and 1, similar to `forget' gates used in LSTMs~\citep{hochreiter1997long}. We then calculate the neuron feedforward activity before the nonlinearity as
\begin{equation}\label{eq:rec_dynamics}
    v_{i}^{\mathrm{ff}}(t) =
    W_i \phi\left(v_{i-1}(t)\right)
     \odot 
    \sigma\left(R_i|r_i(t)|\right)
\end{equation}
where $R_{i}$ refers to the recurrent weight matrix in the $i$-th layer. After applying the effect of the recurrent gating, we re-scale the population activity to have the same overall magnitude as before applying the gating. We thus only change the distribution, but not the total level of activity. At convergence, we learn the recurrent gating weights according to a rule inspired by the feedforward updates from Equation \ref{eq:update_W_ss}

\begin{equation}\label{eq:update_R_i_general}
    \Delta R_{i} =
      \eta  (|r_{i,\mathrm{ss}}| - |\phi(v_{i,\mathrm{ss}}^{\mathrm{ff}})|) |r_{i,\mathrm{ss}}|^T
\end{equation}
where $r_{i,\mathrm{ss}}$ are the target activations of the presynaptic neurons in the same layer. Because our multiplicative gating mechanism affects the magnitude, but not the sign of the activity, we render this inhibition to depend on the magnitude of presynaptic activity. We therefore use absolute values of activity in both the dynamics (Equation \ref{eq:rec_dynamics}) and the update rule (Equation \ref{eq:update_R_i_general}). Like forward weight updates, we normalize recurrent weight updates to zero mean. In contrast to the feedforward weights, however, we only update incoming weights of inactivated neurons (i.e. neurons with activity set to zero by the winner-take-all sparsity mechanism). This lets us simplify the above equation to a Hebbian-like update rule for suppressed neurons:

\begin{equation}\label{eq:update_R_i}
    \Delta R_{i} = 
      -  \eta |\phi(v_{i,\mathrm{ss}}^{\mathrm{ff}})|
      |r_{i,\mathrm{ss}}|^T.
\end{equation}

As a result, we only update incoming recurrent weights for inactive neurons within the target representation, while for active neurons, we only update the incoming feedforward weights. Fig.~\ref{fig1} (dashed lines) summarizes the weight updates. As in standard DFC, we use a simple feedforward pass during test time, for which neither top-down feedback nor lateral recurrent effects are taken into account.

\section{Experiments}\label{sec:results}

To test the CL capabilities of our approach, we train sparse-recurrent DFC on the split-MNIST dataset, according to the domain- and class-IL paradigms \citep{vandeven2019}. Split-MNIST is a simple computer vision CL benchmark in which five pairs of consecutive digits are presented as a sequence of individual supervised learning tasks. In domain-IL, all tasks involve predicting the parity (even/odd) of the input digit, meaning that the output labels stay the same across tasks, but the input data changes. In class-IL, a different class has to be predicted for every digit, so that, across tasks, both the input digits and the class labels change.

\subsection{Performance}\label{sec:performance}

\begin{figure*}
    \centering
    \includegraphics[width=174mm]{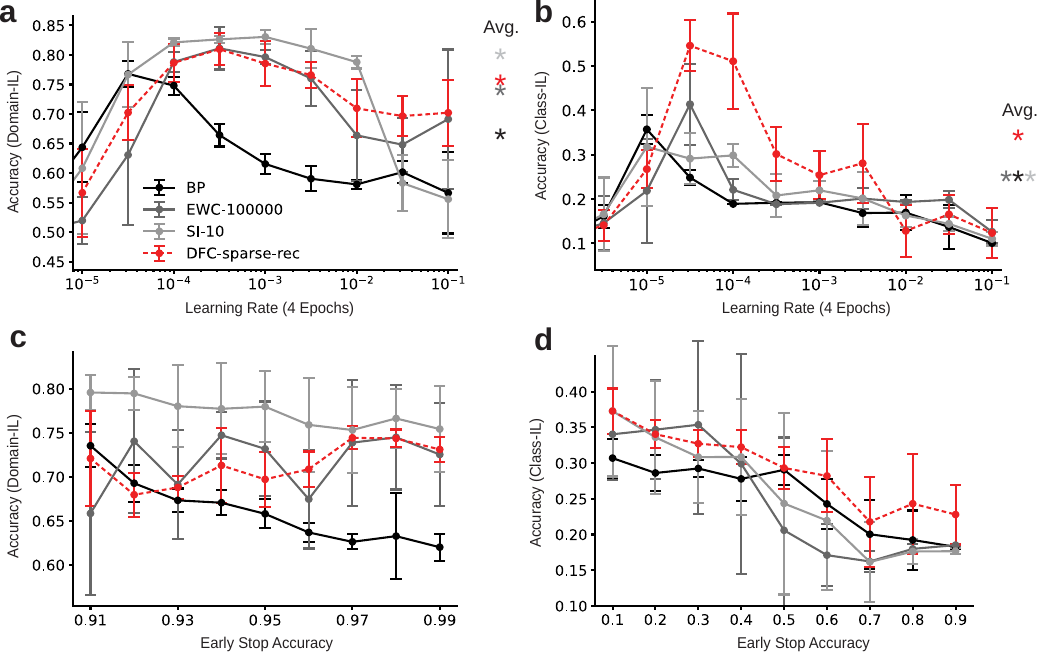}
    \caption{Performance evaluation in split-MNIST for BP, EWC, SI, and DFC-sparse-rec for domain-IL (left column) and class-IL (right column). Error bars represent standard deviations using five random seeds. 
    \textbf{a} Split-MNIST accuracy at the end of training in the domain-IL paradigm on the whole test set (all digits) for a range of LRs. The number of training iterations is fixed at four epochs. Stars indicate average performance on an accuracy-maximizing window of six LRs. \textbf{b} Accuracy of models at the end of training in the class-IL paradigm on the whole test set for every LR. Stars indicate average performance on an accuracy-maximizing window of six LRs. \textbf{c} Accuracy of models at the end of training in the domain-IL paradigm on the whole test set for a range of minimum early stop accuracies. The LR is fixed, and training is stopped at every task once the train accuracy for the current batch reaches the given minimum accuracy value. The maximal number of epochs trained for is 10. \textbf{d} Accuracy of models at end of training in the class-IL paradigm on the whole test set for a range of minimum early stop accuracies}
    \label{fig2}
\end{figure*}

To establish whether sparse-recurrent DFC actually succeeds at CL, we compare its performance against other learning algorithms, namely Synaptic Intelligence (SI), Elastic Weight Consolidation (EWC), as well as standard BP as baseline. Previous studies evaluated models at a fixed learning rate (LR) for a fixed number of epochs \citep{Kirkpatrick2016, vandeven2019}, however, we consider this problematic. Both the LR and the number of epochs can be seen as indicators for how much a network learns, thus pointing to an inherent trade-off between learning the current task well and forgetting previous tasks. Less learning generally leads to less forgetting, while at the same time not allowing the training to converge on the current task. Comparing CL algorithms at a single LR for a fixed number of training samples is problematic for two reasons. First, it does not account for different (model-specific) optimal amounts of training. Second, it fails to capture how robust a CL approach is to more learning, beyond its optimum LR and number of training samples per task. To overcome this issue, we evaluate learning algorithms in two different scenarios.
In the first scenario we fix the number of epochs and vary the LR. In the second scenario we fix the LR and vary the training accuracy that we expect on the current task, before training on the next task, which results in different numbers of batches trained on for different models on different tasks. In both scenarios we cover a wide spectrum between minimizing forgetting, and optimizing the current task.
\subsubsection{LR performance evaluation}
Figures \ref{fig2}a and \ref{fig2}b show performance for a fixed number of training samples across a range of LRs for domain-IL and class-IL, respectively. The initial rise of performance followed by a decay can be explained by the fact that very small LRs (left of the peak) generally prevent sufficient learning while high LRs (right of peak) lead to catastrophic forgetting. These CL performance profiles confirm our initial intuition that choosing a single LR to compare CL methods might lead to overestimating one method over another. We regard good performance in this setting as a function of both peak accuracy and the degree to which accuracy can be maintained once the optimal LR is reached. In domain-IL, sparse-recurrent DFC significantly outperforms BP and achieves a similar performance profile to EWC. Compared to SI, our approach performs worse in terms of peak accuracy, but maintains accuracy over 70\% for a wider range of LRs. In class-IL, sparse-recurrent DFC outperforms all other methods both in peak accuracy and average accuracy.
\subsubsection{Early stop performance evaluation}
Figures \ref{fig2}c and \ref{fig2}d show performance for a fixed LR across a range of early stop accuracies for domain-IL and class-IL, respectively. In domain-IL, sparse-recurrent DFC outperforms BP for almost all minimum accuracies. However, it is most competitive when we train each task to convergence. For training up to very high accuracies, sparse-recurrent DFC is comparable to both EWC and SI. In class-IL sparse-recurrent DFC outperforms all other CL algorithms for the majority of accuracies.

Overall, we conclude that sparse-recurrent DFC represents a competitive CL method that shows a robust performance independent of the amount of learning on each individual task. In the next section, we investigate in more detail the effects on accuracy with respect to the main components of our method: feedback, sparsity and intra-layer recurrency.

\subsection{Integrating feedback signals facilitates CL}
A major difference between standard BP and DFC is that in DFC, the activity of each neuron during training reflects feedforward as well as feedback (error) signals coming from the top-down controller. As a result, target representations $r_{i, \mathrm{ss}}$ are specific to both input and output, with data points exhibiting larger overlaps in active neuron populations if these have similar features \textit{or} the same label.
Fig.~\ref{fig3}a shows that the CL performance is improved across a wide range of LRs if we take into account feedback signals when selecting the remaining active population within the sparse target. We conclude that the feedback signals facilitate the sparsity selection procedures so that more task-specific representations can form. 

\subsection{Sparsity and recurrent gating are required for CL}
\begin{figure*}[h]
    \centering
    \includegraphics[width=174mm]{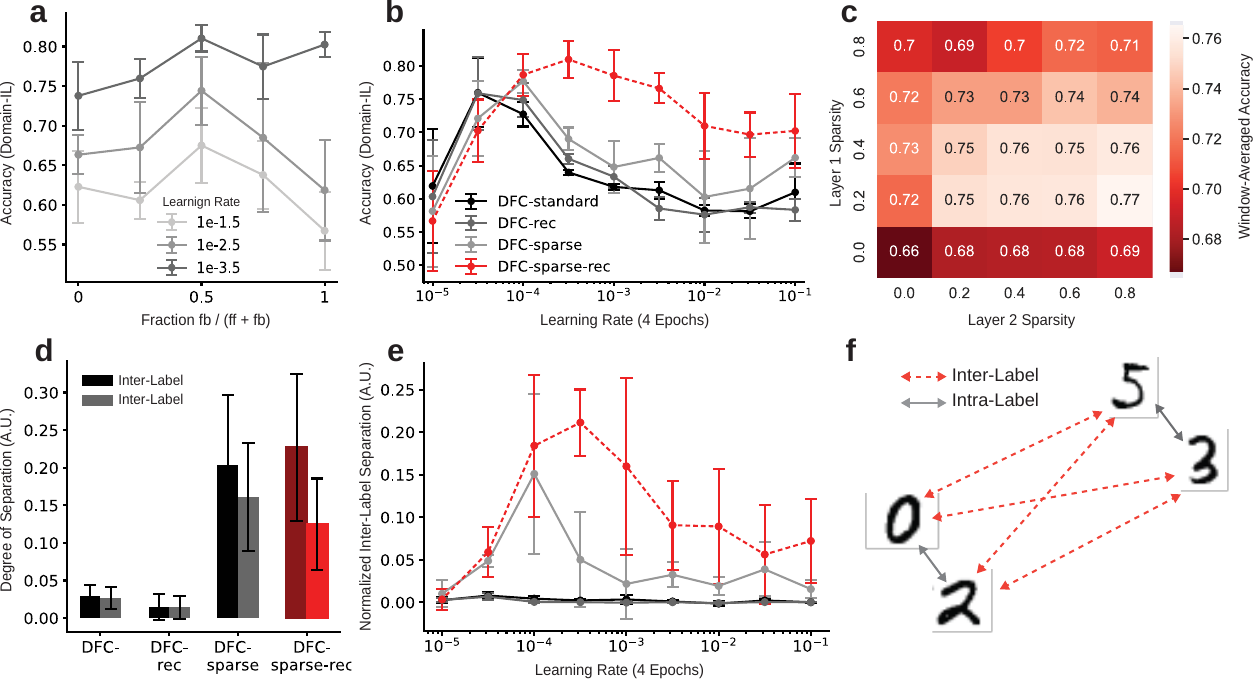}
    \caption{Necessity of sparse-recurrent DFC components and activity separation analysis for domain-IL. Error bars represent standard deviations using five random seeds. \textbf{a} Split-MNIST performance when using varying ratios of feedforward and feedback activity to select the suppressed population for different LRs. The x-axis represents the fraction of feedback activity used for the selection of neurons to be suppressed. A value of 0 means only feedforward activity (ff) is considered, a value of 1 means only feedback (fb) is taken into consideration, and 0.5 corresponds to an equal mix of the two activities. This activity mix is only used for selecting the active neuron population, but the activity flowing through the neurons corresponds to the normal network activity given by Equation \ref{eq:network_dynamics}. \textbf{b} Cross-LR evaluation for all DFC variants. The plot reflects the overall performance on all split-MNIST digits at the end of training. 
    \textbf{c} Cross-LR accuracy for different combinations of hidden layer sparsity levels. The accuracies were aggregated to single numbers by averaging over a contiguous window of six LRs that maximizes average performance, and over five random seeds.
    \textbf{d} Inter- and intra-label separations for DFC variants after all five tasks have been learned. Intra-label separations are calculated for all digit pairs with the same label, inter-label separations for all pairs of digits with different labels. Results are averaged over nine LR values. \textbf{e} Normalized inter-label separation calculated as the difference between inter-label separation and intra-label separation at the end of training across a range of LRs.
    \textbf{f} Visualization of intra- and inter-label distances in the space of activity separation}
    \label{fig3}
    
\end{figure*}

We next investigate whether both sparsity and intra-layer recurrence in the DFC framework are crucial for CL. We compare the accuracy of sparse-recurrent DFC against standard DFC, sparse DFC and recurrent DFC. As opposed to sparse-recurrent DFC, recurrent DFC has no inactivated neurons to constrain the recurrent weight updates to. We thus apply the recurrent weight update rule from Equation \ref{eq:update_R_i_general} to all neurons. Fig.~\ref{fig3}b shows that neither sparsity nor recurrent gating alone significantly alter CL performance across LRs. However, the combination of the two leads to better performance across a wide range of LRs.

Figure \ref{fig3}c shows accuracy as a function of the sparsity parameters $s_{i, \mathrm{ss}}$. For the first hidden layer, a small but non-zero sparsity level yields the best performance while for the second hidden layer, higher sparsity levels work best. This dependence on layer depth is expected, because the early layers of deep networks encode low-level features common to multiple classes and class-selectivity is a disadvantage for these neurons \citep{Morcos2018}, while the later layers encode higher-level features which are more specific to individual classes \citep{zeiler2014visualizing, mahendran2016visualizing}.

\subsection{Aligning sparse, separated representations across tasks facilitates domain-IL}
\label{sec:hyperplane-analysis}
Next, we test whether the combination of sparsity and recurrent gating facilitates CL by reducing representational overlap, in a domain-IL setting. We compute the reduction in overlap (i.e.\ separation) between last hidden layer representations of all pairs of digits, at the end of training. We distinguish between intra-label separation (MNIST digits with the same parity label) and inter-label separation (digits with different parity labels), as shown in Figure \ref{fig3}f. We compute representational separation between digits as

\begin{equation}\label{eq:rep-sep}
s(d_1, d_2) = 1 -  \frac{a_{l}^{d_1}\cdot a_{l}^{d_2}}{\|a_{l}^{d_1}\|\|a_{l}^{d_2}\|}; \qquad a_{l}^d = \sum_{j = 1}^n |r_{l,j}^d|
\end{equation}
where $r_{l,j}^d$ represents the activations in layer $l$ elicited by the $j$'th sample of digit $d$. Fig.~\ref{fig3}d shows the averages of inter- and intra-label representational separations for DFC variants. Interestingly, sparse DFC shows high representational separation, but does not yield significantly higher accuracies compared to standard DFC or BP. This suggests that overall increases in representational separation alone do not account for performance improvements that we observe in Fig.~\ref{fig3}b.

To better understand this result, we next devise a new measure of separation, which we term normalized inter-label separation and that is defined as the average difference between inter-label separation and intra-label separation. Fig.~\ref{fig3}e shows this separation metric over a wide range of LRs. For the LRs where sparse-recurrent DFC yields higher normalized inter-label separation, we also observe better CL performance (compare to Fig.~\ref{fig3}b), suggesting that the relative degree of digit representational overlap can explain the CL performance profile that we observe for sparse-recurrent DFC. This indicates that sparse-recurrent DFC facilitates domain-IL performance by representing even and odd digits in two partially separated neuron populations that are reused across tasks. As a first result, we conclude that, although sparsity is necessary to create non-overlapping representations, sparsity alone is not sufficient for aligning these across tasks. Such alignment, however, seems beneficial for domain-IL, where several digits are represented by the same label. We next investigate how recurrent gating helps to learn representations that are compatible across tasks.

\begin{figure*}[h]
    \centerline{\includegraphics[width=174mm]{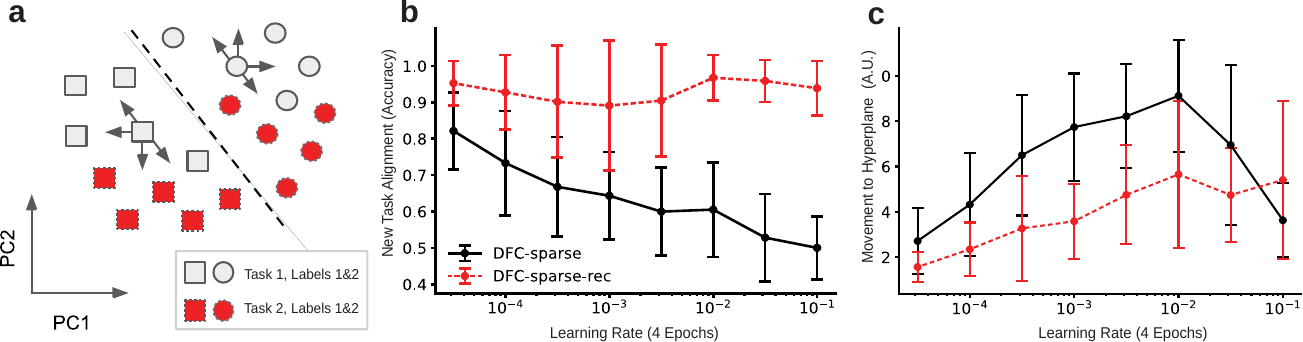}}
    \caption{Effects of recurrent gating on last hidden layer targets $r_{L-1,\mathrm{ss}}$ and feedforward activations $\phi(v_{L-1}^{\mathrm{ff}})$ during learning. Error bars represent standard deviations using five random seeds. \textbf{a} Schematic of task 1 and 2 representations with respect to the hyperplane (dashed line) dividing task 1 target activations (grey) according to their label. This diagram illustrates two things: First, the new target representations align with the previously learned hyperplane in terms of label separation (supported by b). In other words, the hyperplane that separates task 1 targets also separates task 2 targets. Second, task 1 representations generally move less towards the separating hyperplane as subsequent tasks are learned (supported by c). This is represented by the arrows. \textbf{b} Alignment of new task target activations with previous hyperplanes. This is measured as the fraction of initial target representations ($r_{L-1,\mathrm{ss}}$, before learning task $i$), of the new task $i$ that are correctly separated according to the hyperplane learned on the previous task $i-1$. \textbf{c} Movement of feedforward activations $\phi(v_{L-1}^\mathrm{ff}$) of previous tasks towards the hyperplane after learning subsequent tasks, normalized by movement in all directions
    }
    \label{fig4}
\end{figure*}

The final hidden layer of a network has to learn representations of the input that are linearly separable by its readout weights. One possible way to prevent CF is to ensure two things.  \textbf{Condition 1}: The hyperplane separating representations of different labels (implemented in the network by the readout layer) needs to stay the same, or similar across tasks. \textbf{Condition 2}: Data points represented in the final hidden layer need to stay on the same side of the classification hyperplane that was initially learned as we train on subsequent tasks. We measure feedforward activations $\phi(v_{L-1}^{\mathrm{ff}})$ (no recurrent gating) and target activations $r_{L,\mathrm{ss}}$ (including effects of controller and recurrent gating) to test whether recurrent gating helps to achieve this. Regarding condition 1, Fig.~\ref{fig4}b shows that, if we classify target activations at training onset of a new task according to the previously learned separation boundary, sparse-recurrent DFC consistently yields higher classification accuracies than sparse DFC. This suggests that lateral connections regularize new target activations such that they better align with previously learned task boundaries. This idea is illustrated in Fig.~\ref{fig4}a, showing that target activations of the second task are separated by the same hyperplane that divides targets of the first task. Regarding condition 2, we measure the direction of movement of feedforward activations from the beginning to the end of training. We next quantify how much the data points move towards the initially learned separation boundary. Fig.~\ref{fig4}c suggests that sparse-recurrent DFC reduces the movement towards the previous decision boundary compared to sparse DFC. Taken together, our results suggest that recurrent gating helps fulfill both conditions. For more details on the calculation of these metrics involving hyperplanes, see Appendix \ref{app:hyperplane-calc}.

\subsection{Learning within separate sub-spaces facilitates class-IL}
\begin{figure*}[h]
    \centerline{\includegraphics[width=174mm]{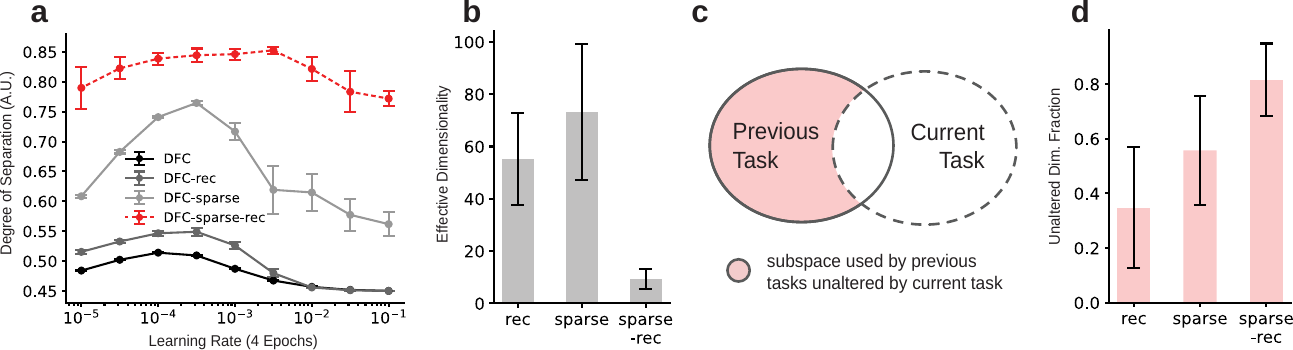}}
    \caption{Last hidden layer target activation ($r_{L-1,\mathrm{ss}}$ belonging to task $i$, after learning task $i$) analysis for class-IL. Error bars represent standard deviations over five random seeds. 
    \textbf{a} Representational separation (Equation \ref{eq:rep-sep}) between pairs of digits for DFC variants for a range of LRs.
    \textbf{b} Effective dimensionality \citep{roy2007effective} of targets averaged over tasks and random seeds for DFC variants for $\text{LR}=0.001$.
    \textbf{c} Visualization of the 'unaltered dimensionality fraction' $\gamma$ measure described in Appendix \ref{app:unaltered-subspace-metric}. The left and right ellipses represent the subspace used by the first $i-1$ tasks, and by task $i$, respectively. $\gamma$ quantifies the dimensionality of the colored area as a fraction of the dimensionality of the area of the left ellipse.
    \textbf{d} Unaltered dimensionality fraction $\gamma$ (described in Equation \ref{eq:unaltered-fraction} from  Appendix \ref{app:unaltered-subspace-metric} and visualized in subplot c) for DFC variants.
    }
    \label{fig5}
\end{figure*}

One possible strategy to address class-IL is to enforce sparse, non-overlapping representations of different digits, thereby preventing interfering weight updates between classes. To test whether sparse-recurrent DFC utilizes this strategy, we record target activities of different digits after they are first learned and measure the representational overlap of all pairs of digits using Equation \ref{eq:rep-sep}. Figure \ref{fig5}a shows that, while sparse DFC leads to some increase in representational separation, sparse-recurrent DFC maximizes separation across all LRs compared to other DFC variants. These results are consistent with our initial idea of reduced representational overlap facilitating CL. Intuitively, if different neurons are used for different tasks, weights of neurons that were important in early tasks are less likely to be changed. Similar to domain-IL, sparsity in class-IL can thus be seen as a necessary condition for the formation of non-overlapping representations.

To gain a better understanding of why recurrent gating helps to increase representational separation in class-IL, we next analyze its effect on altering the dimensionality of targets. Figure \ref{fig5}b shows the effective dimensionality \citep{roy2007effective} of the target activations of different tasks after learning for recurrent DFC, sparse DFC and sparse-recurrent DFC. The results suggest that the combination of sparsity and recurrent gating leads to a significant decrease in effective dimensionality of the target activations. This led us to hypothesize that representations learned for a new task are less likely to affect dimensions that were important for previous tasks. To investigate if recurrent gating leads to a reduction in reuse of previously learned subspaces, we compute the fraction of the effective dimensionality used by previous tasks that is altered by the current task (Fig. \ref{fig5}c). For more details on the calculation of this metric, see Appendix \ref{app:unaltered-subspace-metric}. Figure \ref{fig5}d validates our hypothesis that recurrent gating reduces the fraction of dimensions that are altered by new tasks, thus reducing the extent to which new weight updates interfere with parameters important for previous tasks.

\section{Discussion}
In summary, we have presented a new, bio-inspired, task-free CL approach that yields competitive performance compared to other CL methods on a simple computer vision benchmark. To restrict learning to a reduced set of task-specific parameters, our method (sparse-recurrent DFC) integrates feedforward and feedback information to constrain activity to a sub-population of neurons. In addition to being more biologically plausible, we show that including top-down signals is beneficial for CL. Our results are consistent with the idea that sparsity is a requirement for reducing representational overlap, but suggest that sparsity alone is insufficient for protecting previously learned model parameters. We show that intra-layer recurrent connections, when combined with sparsity, facilitate the protection of old task representations, leading to competitive CL performance of DFC on split-MNIST. For both domain- and class-IL, recurrent gating in combination with sparsity restricts learning to low-dimensional subspaces. In domain-IL the same subspace consisting of two separated neuron populations is shared across tasks; in class-IL learning is restricted to multiple distinct subspaces. 

From a neuroscience perspective, our findings might allow experimental researchers to derive new hypotheses about how the brain minimizes CF. One prediction of our sparse-recurrent DFC network is that intra-layer recurrent connections are only critical during learning but not inference, since we only use recurrence at training time. Although this is surprising, there are data suggesting that biological brains do this as well. \cite{van1998face} argue that, given the short response time in face recognition tasks, neurons do not have the time to emit much more than one spike at each processing stage. This would imply that initial inference can happen before recurrence takes effect. Based on our work, neuroscientists could, for example, manipulate recurrent communication within cortical hierarchies, to test if an animal's ability to perform inference or to learn multiple tasks sequentially is affected. 

From a machine learning perspective, our new method is relevant because it is based on a novel set of working principles to achieve CL. As sparse-recurrent DFC naturally infers non-overlapping representations and thus non-interfering parameter updates, it does not require any task boundaries or task information either during training or testing. While other task-free CL methods exist and achieve competitive performance, they are not exclusively based on specialized weight update rules, as they use either data replay or expanding architectures. The only exception we could find is limited to binary networks \citep{Laborieux2021}. Moreover, in future work, our approach could be combined with other task-free CL methods (replay and non-replay-based) which might lead to even better CL performances. Although the current implementation of sparse-recurrent DFC is computationally less efficient when compared to standard CL algorithms running on GPUs, DFC is ideally suited for a neuromorphic hardware implementation that might be more energy-efficient. 

Overall, our work showcases the idea of adopting biological principles of neural computation and learning to derive new CL methods that not only perform significantly better than BP, but also show performance comparable to existing CL algorithms. 

\section*{Declarations}
\footnotesize
\subsection*{Funding}
This work was supported by the Swiss National Science Foundation (B.F.G. CRSII5-173721 and 315230\_189251), ETH project funding (B.F.G. ETH-20 19-01), the Human Frontiers Science Program (RGY0072/2019) and funding from the Swiss Data Science Center (B.F.G, C17-18). P.V.A. was supported by an ETH Zürich Postdoc fellowship. M.S. was supported by an ETH AI Center postdoctoral fellowship.

\subsection*{Competing interests}
The authors have no relevant financial or non-financial interests to disclose.

\subsection*{Author Contributions}
F.L., B.F.G., P.A. conceptualized the project idea. F.L. and M.S. carried out all CL experiments and simulations. F.L., B.F.G., P.A. and M.S. wrote the manuscript. F.L. and B.F.G. made the figures and plots.

\subsection*{Availability of code}
The implementation of all of the models, training pipeline and analysis code are available on GitHub\footnote{https://github.com/pennfranc/bio-inspired-continual-learning} and archived\footnote{https://doi.org/10.5281/zenodo.7414720}. Additionally, a modified version of the \textit{hypnettorch} library adapted from \cite{oshg2019hypercl} used to load split-MNIST data is likewise available on GitHub\footnote{https://github.com/pennfranc/hypnettorch} and archived\footnote{https://doi.org/10.5281/zenodo.7361495}.
\label{app:data-materials}
The implementations of EWC and SI were adapted from \cite{Hsu2018} to be compatible with our codebase.

\bibliographystyle{plainnat}
\bibliography{bibliography}

\clearpage
\appendices
\normalsize

\section{Accuracy vs. tasks learned}
\label{app:acc-vs-tasks}
In Section \ref{sec:performance} we evaluate the performance of models on split-MNIST by recording the test accuracies after training on all tasks. To analyze how cumulative performance develops as more tasks are learned, we plot the mean accuracy of the first $i$ tasks after training on task $i$ (Figure 6). Each model was evaluated with its optimal LR for 4 epochs. Curves that start with low accuracies for task 1 can be explained by the fact that choosing an LR that leads to convergence on task 1 is not optimal for the final accuracy on all tasks. Moreover the increase of cumulative accuracy for task 4 in domain-IL can be attributed to the similarity of the digit pairs 0/1 and 6/7.

\begin{figure}[htbp]
    \centering
    \includegraphics[width=80mm]{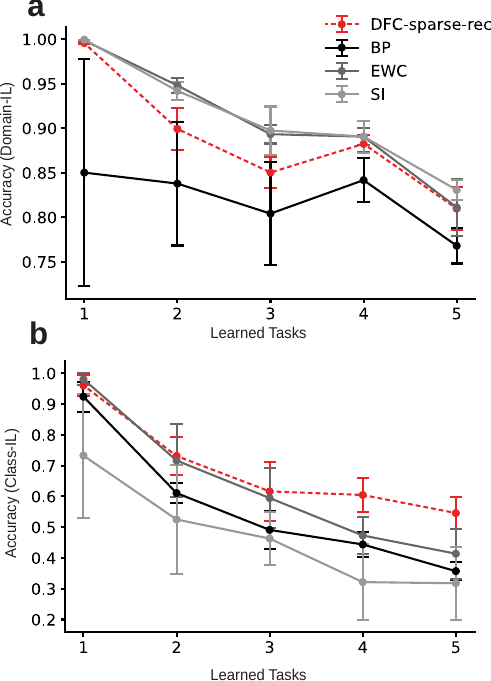}
    \caption{Average accuracy of first $i$ tasks after training on $i$'th task. LRs were chosen for each model individually to maximize performance. All models were trained for four epochs.
    \textbf{a} Cumulative domain-IL accuracies for first $i$ tasks on test set.
    \textbf{b} Cumulative class-IL accuracies for first $i$ tasks on test set
    }\label{fig6}
\end{figure}

\section{Hyperparameters}
\label{app:hyperparameters}

Our approach for choosing hyperparameters in sparse-recurrent DFC is to start with a configuration that is optimized to solve normal MNIST classification (non-CL) \citep{Meulemans2022}, and to leave all existing parameters unaltered for split-MNIST. Adding sparsity and recurrent gating introduces layer-wise sparsity levels and recurrent learning rate, respectively, as new hyperparameters. These new hyperparameters were tuned separately for domain-IL and class-IL. For EWC and SI, we tuned the regularization coefficient. The overarching principle here is that we only tune hyperparameters specifically associated with solving CL. Table \ref{hyperparameter-table-1} shows all tuned hyperparameters, as well as the activation function (which was not tuned). Table \ref{hyperparameter-table-2} shows the remaining hyperparameters shared by all models.

\begin{table}[h]
\caption{\label{hyperparameter-table-1}Model-specific hyperparameters. Except for the activation function, all of these hyperparameters were tuned for performing well on split-MNIST in a cross-LR evaluation paradigm. The three numbers in the sparsity level rows correspond to the 2 hidden layers and the output layer respectively. The effect of sparsity in the output layer is solely to freeze weights of inactive neurons during training for wrong labels}

\centering
\begin{tabular}{|l|l|l|l|l|}
\hline
                             & \textbf{DFC} & \textbf{BP} & \textbf{EWC} & \textbf{SI} \\ \hline
sparsity levels domain-IL & 0.4,0.8,0.5  & -           & -            & -           \\ \hline
sparsity levels class-IL  & 0.2,0.8,0.0  & -           & -            & -           \\ \hline
recurrent learning rate      & 40           & -           & -            & -           \\ \hline
reg. coefficient domain-IL   & -            & -           & 200          & 10          \\ \hline
reg. coefficient class-IL    & -            & -           & 200          & 100         \\ \hline
activation function          & tanh         & relu        & relu         & relu        \\ \hline
\end{tabular}
\end{table}

\begin{table}[h]
\caption{\label{hyperparameter-table-2}Hyperparameters shared between all used models}
\centering

\begin{tabular}{|l|l|}
\hline                                         & \textbf{All models} \\ \hline
\# hidden layers                         & 2                   \\ \hline
hidden layer sizes domain-IL             & 20,20               \\ \hline
hidden layer sizes class-IL              & 200,200             \\ \hline
learning rate (outside of LR evaluation) & 0.001               \\ \hline
batch size                               & 512                 \\ \hline
epochs                                   & 4                   \\ \hline
optimizer                                & adam                \\ \hline
forward weight initialization            & xavier              \\ \hline
\end{tabular}
\end{table}

\section{Hyperplane metrics}
\label{app:hyperplane-calc}
In Section \ref{sec:hyperplane-analysis} we compute two quantities that involve the use of hyperplanes dividing datapoints into two classes, as per the domain-IL setup \citep{vandeven2019}. In both cases we obtain the separation hyperplane by fitting a logistic regression model to a set of target activations of the last hidden layer $\{r_{L-1}^{k, j}\}_{k\in t_{i}}$, where $t_{i}$ refers to a set of indices of datapoints belonging to task $i$. $r_{L-1}^{k, j}$ represents the last hidden layer target activations induced by datapoint $k$ after that network has been trained on task $j$. Let $h_{i,j}$ denote the hyperplane obtained by fitting a logistic regression model to classify $\{r_{L-1}^{k, j}\}_{k\in t_{i}}$ according to the domain-IL class labels. We use an L1 penalty for the logistic regression model to encourage sparse hyperplanes, otherwise we use the default parameters from the sci-kit learn library \citep{scikit-learn}.

\subsection{Hyperplane alignment}
Here we measure the extent to which $\{r_{L-1}^{k, i-1}\}_{k\in t_{i}}$ are correctly separated by $h_{i-1,i-1}$, that is how well a hyperplane from a previously learned task $i-1$ divides targets of new tasks $i$, before the network has been fit on the new task. If we represent classification accuracy of $h_{i,j}$ on $\{r_{L-1}^{k, u}\}_{k\in t_{v}}$ ($i$, $j$ and $u$, $v$ representing arbitrary task indices) as $h_{i,j}(\{r_{L-1}^{k, u}\}_{k\in t_{v}})$, then the hyperplane alignment metric $\alpha$ is given by Equation \ref{eq:hyperplane-alignment}.

\begin{equation}
\label{eq:hyperplane-alignment}
\alpha = \frac{1}{4}\sum_{i=2}^5h_{i-1,i-1}(\{r_{L-1}^{k, i-1}\}_{k\in t_{i}})
\end{equation}
$\alpha$ values are further averaged over 5 random seeds.

\subsection{Movement towards hyperplane}
For this metric, we consider distances travelled of feedforward activations, which we would normally refer to as $\phi(v_i^\mathrm{ff})$. But because we are running out of space for superscripts, we will refer to $\tilde{r}_{L-1}^{k, j}$ as the last hidden layer feedforward activations induced by datapoint $k$ after that network has been trained on task $j$. Please note, however, that $h_{i,j}$ is still computed as before, using target activations (including controller and recurrent effects). We quantify the distance of feedforward activations traveled from when they are first learned, to when task $5$ training has been finished, with respect to the initially learned hyperplane. More precisely, for all task indices $i$, we compute the difference of the projections of $\{\tilde{r}_{L-1}^{k, i}\}_{k\in t_{i}}$ and $\{\tilde{r}_{L-1}^{k, i-1}\}_{k\in t_{5}}$ on the normal of $h_{i,i}$, which we denote as $n_{i,i}$.  Let $T_{i,j}^c$ denote the matrix that contains as rows all elements of $\{\tilde{r}_{L-1}^{k, j}\}_{k\in t_{i}}$ which have $c$ as their correct class label, where $c \in \{0,1\}$. From these matrices we can compute the L1 distances traveled by datapoints with class $c$ from task $i$ projected onto the hyperplane normal $n_{i,i}$ as seen in Equation \ref{eq:hyperplane-travel-dist}.
\begin{equation}
\label{eq:hyperplane-travel-dist}
\tilde{d}_i^c = (-1)^c \cdot (T_{i,5}^c - T_{i,i}^c)n_{i,i}
\end{equation}
The $(-1)^c$ factor is important to ensure inverted signs of travelled distances in the two classes. We need this because directions towards the hyperplane for one class are directions away from the hyperplane for the other. Because we only want to quantify distance traveled towards the hyperplane direction, and not away from it, we clip the distance vectors to only have positive values.
\begin{equation}
d_i^c = clip(\tilde{d}_i^c, 0, \infty)
\end{equation}
Finally, we obtain the mean normalized movement towards the hyperplane of activations from task $i$ by dividing the average distance traveled towards $h_{i,i}$ by the average absolute distance travelled in any principal direction, as shown in Equation \ref{eq:normalized-hyperplane-movement}.

\begin{equation}
\label{eq:normalized-hyperplane-movement}
\beta_i = \frac{\langle d_i^c\rangle_{c\in\{0,1\}}}{\frac{1}{2}\langle|T_{i,5}^{0,1} - T_{i,i}^{0,1}|\rangle}
\end{equation}

We need to divide the normalizing factor in the denominator by 2 because we are technically averaging over twice as many directions as there are matrix entries. This is because we consider both positive and negative directions for each principle dimension. The $\beta_i$ values are averaged over tasks $i$ and 5 random seeds.

\section{Fraction of unaltered subspace}
\label{app:unaltered-subspace-metric}
With the unaltered subspace metric $\gamma$ we attempt to approximate the idea of the fraction of dimensions used by previous tasks that are left unaltered by the current task, as visualized by Figure \ref{fig5}c. We reuse the notation from the previous section, where $\{r_{L-1}^{k, j}\}_{k\in t_{i}}$ refers to the set of target activations $r_{L,\mathrm{ss}}$ elicited by datapoints of task $i$ upon learning task $j$. To quantify the dimensionality of a set of neural activity vectors of a given layer, we utilize the effective rank metric proposed by \cite{roy2007effective}. The effective rank of a matrix $A$ with positive singular values $\sigma_1, \geq \sigma_2 \geq ... \geq \sigma_Q$ is calculated using Shannon entropy $H$ as shown in Equations \ref{eq:singular-value-distribution} and \ref{eq:effective-rank}.

\begin{align*}
 p_k &= \frac{\sigma_k}{\sum_{k=1}^Q|\sigma_k|} \numberthis \label{eq:singular-value-distribution} \\
 \\
 \text{erank}(A) &= exp(H(p_1, ..., p_Q)) \numberthis \label{eq:effective-rank}
\end{align*}

We compute the effective rank of the matrix containing activity vectors as rows to quantify the effective dimensionality of the representations. We calculate the effective dimensionality of previously learned tasks (up to but without task $i$), the current task $i$, and the combination of previous tasks and the current task as shown in Equations \ref{eq:dim-prev}, \ref{eq:dim-curr}, \ref{eq:dim-cum}, respectively.

\begin{align*}
\text{dim}_{prev}(i) &= \text{erank}(
\{r_{L-1}^{k, j}\}_{k\in \bigcup_{l=1}^{i-1}t_l}
) \numberthis \label{eq:dim-prev} \\
\text{dim}_{curr}(i) &= \text{erank}(r^{t_i}) \numberthis \label{eq:dim-curr} \\
\text{dim}_{cum}(i) &= \max(\text{erank}([r^{t_1}, ..., r^{t_i}]), \\
&\:\:\:\:\:\:\:\:\:\:\:\:\:\:\:\:\: \text{dim}_{prev}(i), \text{dim}_{curr}(i)) \numberthis \label{eq:dim-cum}
\end{align*}

Effective rank as a function of sets of target activations does not guarantee monotonicity, which means that the effective rank of a subset of targets can be larger than the effective rank of the superset. To avoid invalid fractions, we guarantee monotonicity between previous, current and cumulative dimensionality by making sure $\text{dim}_{cum}$ is at least as big as $\text{dim}_{prev}$ and $\text{dim}_{curr}$. If we subtract the cumulative dimensionality from the sum of the previous and the current one, we get the intersection of the two, i.e. the dimensionality that is affected by the current task. To quantify the unaltered fraction of previous dimensionality $\gamma$, we subtract the fraction of the intersection divided by the previous dimensionality from 1 as shown in Equations \ref{eq:unaltered-fraction}.

\begin{equation}
\label{eq:unaltered-fraction}
\gamma = 1 - \frac{\text{dim}_{prev}(i) + \text{dim}_{curr}(i) - \text{dim}_{cum}(i)}{\text{dim}_{prev}(i)}
\end{equation}

\end{document}